\documentclass[11pt]{article}
\usepackage{coling2020}
\usepackage{times}
\usepackage[hyphens,spaces]{url}
\usepackage{latexsym}
\usepackage{graphicx}
\usepackage{subcaption}
\graphicspath{ {./images/} }
\usepackage{todonotes}

\usepackage{fancyvrb}
\usepackage{fancyhdr}
\usepackage{lipsum}

\colingfinalcopy 

\title{Distant Reading of the German Coalition Deal:\\ Recognizing Policy Positions with BERT-based Text Classification}
\author{Michael Zylla$^1$, Thomas Haider$^{1,2}$ \\
$^{1}$University of Göttingen, Germany \\
$^{2}$Max Planck Institute for Empirical Aesthetics, Frankfurt\\
  michael.zylla@stud.uni-goettingen.de, thomas.haider@uni-goettingen.de}

\begin{document}

\maketitle

\section{Motivation}
In postwar Germany, the federal government is usually formed by several political parties \cite[p.~97]{schmidt_2007}.\footnote{This paper was originally presented at the \textit{International Digital Humanities 2022 conference} in Tokyo and published via the book of abstracts: \url{https://dh2022.adho.org/}} 
Over the past 16 years, these government coalitions were led by the Christian Democratic parliamentary group (CDU/CSU), most recently in cooperation with the Social Democratic Party (SPD), which, following the federal election in 2021, was unwilling to negotiate with their former partner, calling for new alliances to achieve a majority in parliament. Finally, the leaders of the Free Democratic Party (FDP), the Greens and SPD, despite mixed support from the party bases, signed a coalition agreement. Some journalists even regarded the FDP, which gained access to two key ministries, the secret winner of the negotiations \cite{smallestparties2021}, also because the Greens did not see some of their desired climate change policies implemented \cite{abstimmungen2021}.

In this research, we are interested in how the coalition agreement was assembled regarding the individual party contributions. To that end, we utilize methods from Natural Language Processing, which have seen widespread adoption in political science \cite{wilkerson_casas,manifesto_2016,Rauh2015,wordfish}. Specifically, we carry out a text classification task with transformer models, based on paragraphs from the party manifestos, and use the resulting model to characterize the coalition deal.

\section{Data}
\begin{figure}[htbp]
    \centering
    \includegraphics[width=0.7\linewidth]{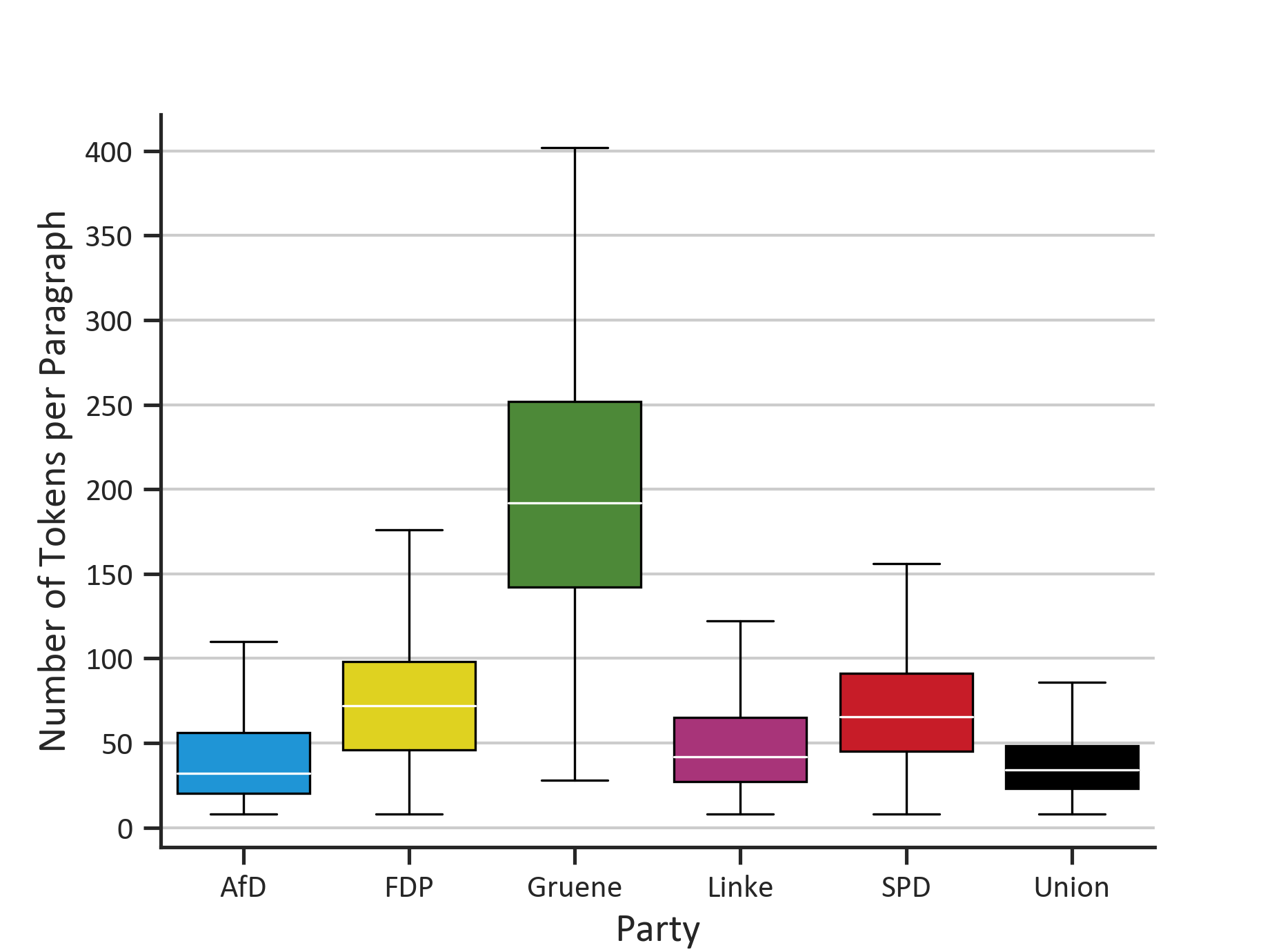}
    \caption{Paragraph length in manifestos}
    \label{fig:boxplots}
\end{figure}

Our data consist of the election manifestos from 2021 of the six parliamentary parties, namely Alternative for Germany (AfD), FDP, Greens (Grüne), Left (Linke), SPD, and CDU/CSU (Union), and also the final coalition deal. We converted the original PDFs to plaintext, removed the tables of contents, cleaned the texts from formatting artefacts, and segmented the documents into individual paragraphs. As seen in Table \ref{tab:data} and Figure \ref{fig:boxplots}, both document and paragraph length vary widely. The manifestos of AfD and SPD in particular are fairly short, when compared to the Greens and the Left.

\begin{table}[htbp]
    \centering
    \begin{tabular}[]{|l|c|c|}
    \hline
    Party & Document Length in Tokens & Number of Paragraphs \\
    \hline
    AfD     &  28,171 & 674 \\
    FDP &  37,710 & 493 \\
    Grüne &  74,757 & 369 \\
    Linke &  76,308 & 1,473 \\
    SPD &  26,553 & 386 \\
    Union &   46,669 & 1,228 \\
    \hline
    \end{tabular}
    \caption{Size of German party manifestos.}
    \label{tab:data}
\end{table}

\section{Experiments}
To investigate the composition of the coalition deal, we trained German BERT \cite{devlin-etal-2019-bert} models on a text classification task with the paragraphs of the party manifestos. We test two models: (1) A classification with all six parties, and (2) a classification with the three coalition partners. We examine the difficulty to classify individual parties, how they are misclassified for each other with confusion matrices, and how confident the classifier is w.r.t. certain paragraphs with a softmax layer (e.g., a paragraph could be assigned 50\% SPD, 30\% Greens and 20\% FDP). Finally, we apply the three-party model to the coalition deal to analyze its composition. 

\section{Results}
\begin{figure}[htbp]
    \begin{minipage}{.5\textwidth}
        \centering
        \includegraphics[width=1\linewidth]{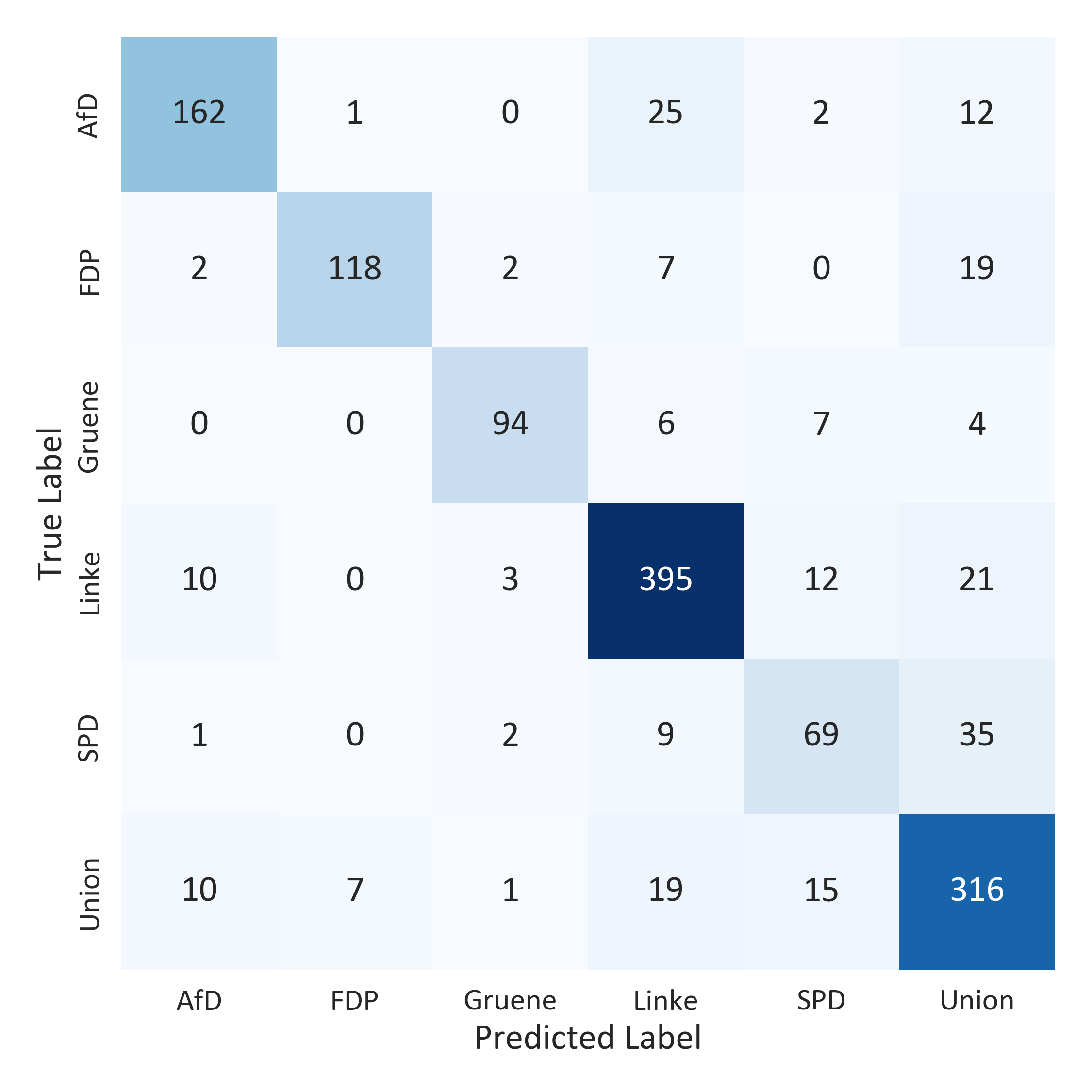}
        \caption{Confusion Matrix of First model}
        \label{fig:confusionmatrix1}
    \end{minipage}%
        \begin{minipage}{.5\textwidth}
        \centering
        \includegraphics[width=1\linewidth]{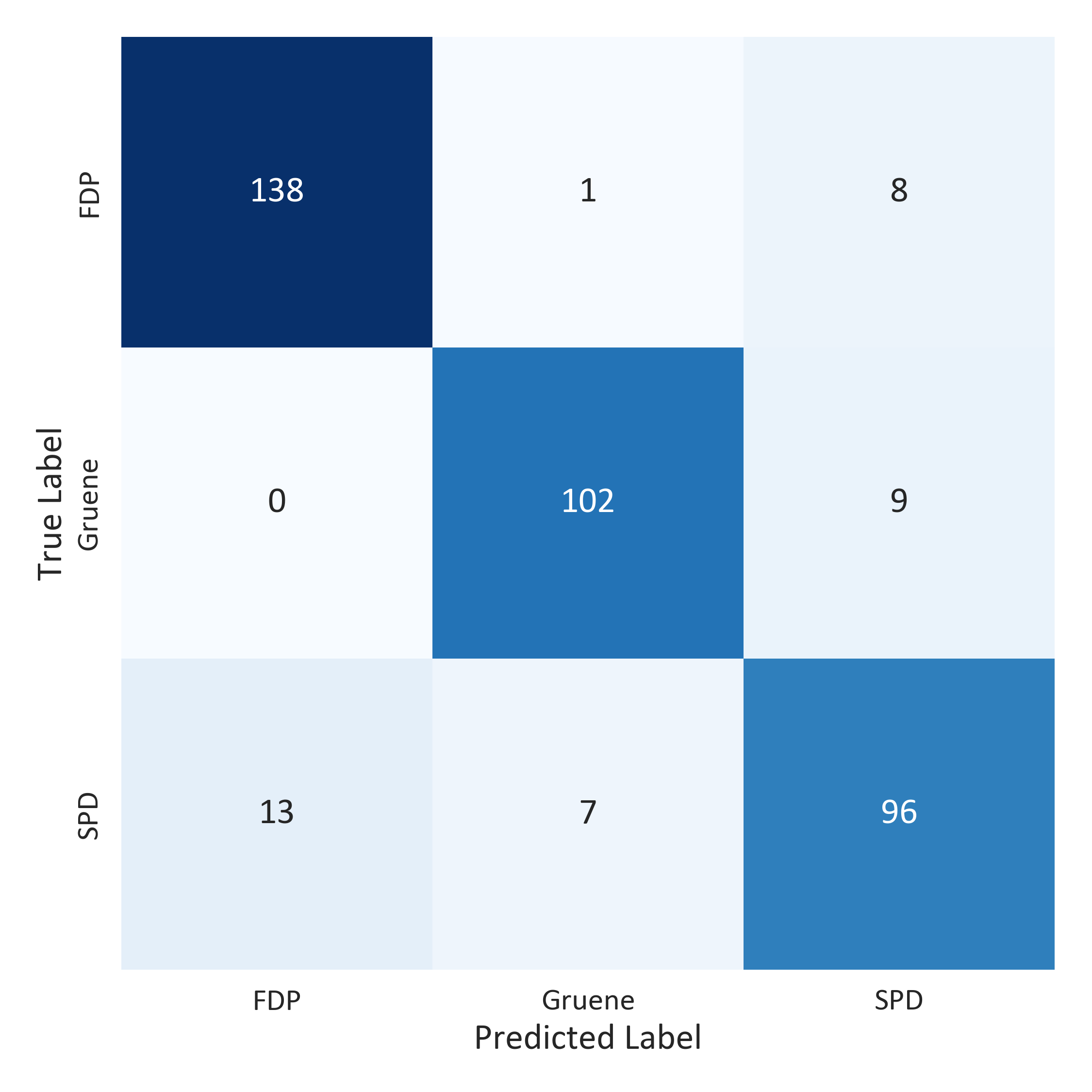}
        \caption{Confusion Matrix of Second model}
        \label{fig:confusionmatrix2}
    \end{minipage}
\end{figure}

We find that the six-class model has markedly more problems recognizing SPD and Union (see Table \ref{tab:fscores_full}). Furthermore, SPD paragraphs are often misclassified as Union, while the inverse is not as frequent (see Figure \ref{fig:confusionmatrix1}). This might be because both are Germany’s largest catch-all-parties, with the other parties having a more distinct vocabulary. Altogether, much of the models’ confusion falls in line with overlapping political positions, e.g., paragraphs from FDP and AfD are misclassified as Union, while the Greens are never mistaken for the former two. On the other hand, Linke and AfD may be mistaken for each other due to their oppositional language. Thus, it is an open question to what extent the model makes decisions based on policy or language.

\begin{table}[htbp]
    \noindent\begin{minipage}[c]{.5\linewidth}
        \centering
        \resizebox{1\linewidth}{!}{
        \begin{tabular}{|l|c|c|c|c|}
            \hline
             & precision & recall & f1-score & support \\
            \hline
            AfD & 0.8757 & 0.8020 & 0.8372 & 202 \\
            FDP & 0.9365 & 0.7973 & 0.8613 & 148 \\
            Gruene & 0.9135 & 0.8559 & 0.8837 & 111 \\
            Linke & 0.8565 & 0.8934 & 0.8746 & 441 \\
            SPD & 0.6699 & 0.5948 & 0.6301 & 116 \\
            Union & 0.7770 & 0.8614 & 0.8170 & 368 \\
            accuracy &  &  & 0.8333 & 1386 \\
            macro avg & 0.8382 & 0.8008 & 0.8173 & 1386 \\
            weighted avg & 0.8357 & 0.8333 & 0.8327 & 1386 \\
            \hline
        \end{tabular}
        }
        \caption{Evaluation of First model}
        \label{tab:fscores_full}
    \end{minipage}%
    \begin{minipage}[c]{.5\linewidth}
        \centering
        \resizebox{1\linewidth}{!}{
        \begin{tabular}{|l|c|c|c|c|}
            \hline
             & precision & recall & f1-score & support \\
            \hline
            FDP & 0.9026 & 0.9456 & 0.9236 & 147 \\
            Gruene & 0.9189 & 0.9189 & 0.9189 & 111 \\
            SPD & 0.8532 & 0.8017 & 0.8267 & 116 \\
            accuracy &  &  & 0.8930 & 374 \\
            macro avg & 0.8916 & 0.8887 & 0.8897 & 374 \\
            weighted avg & 0.8921 & 0.8930 & 0.8921 & 374 \\
            \hline
        \end{tabular}
         }
        \caption{Evaluation of Second model}
        \label{tab:fscores_ampel}
    \end{minipage}%
\end{table}

The three-class model achieves better classification results (see Table \ref{tab:fscores_ampel}), which is not surprising, since the task is easier with less parties to choose from. However, SPD paragraphs are still harder to predict.

As can be seen in the confusion matrix (Figure \ref{fig:confusionmatrix2}), the model can reliably distinguish FDP and Greens, but both are harder to distinguish from SPD. This, again, might be explained by the SPD’s claim to being a catch-all party.

\begin{figure}[htbp]
    \centering
    \includegraphics[width=0.65\linewidth]{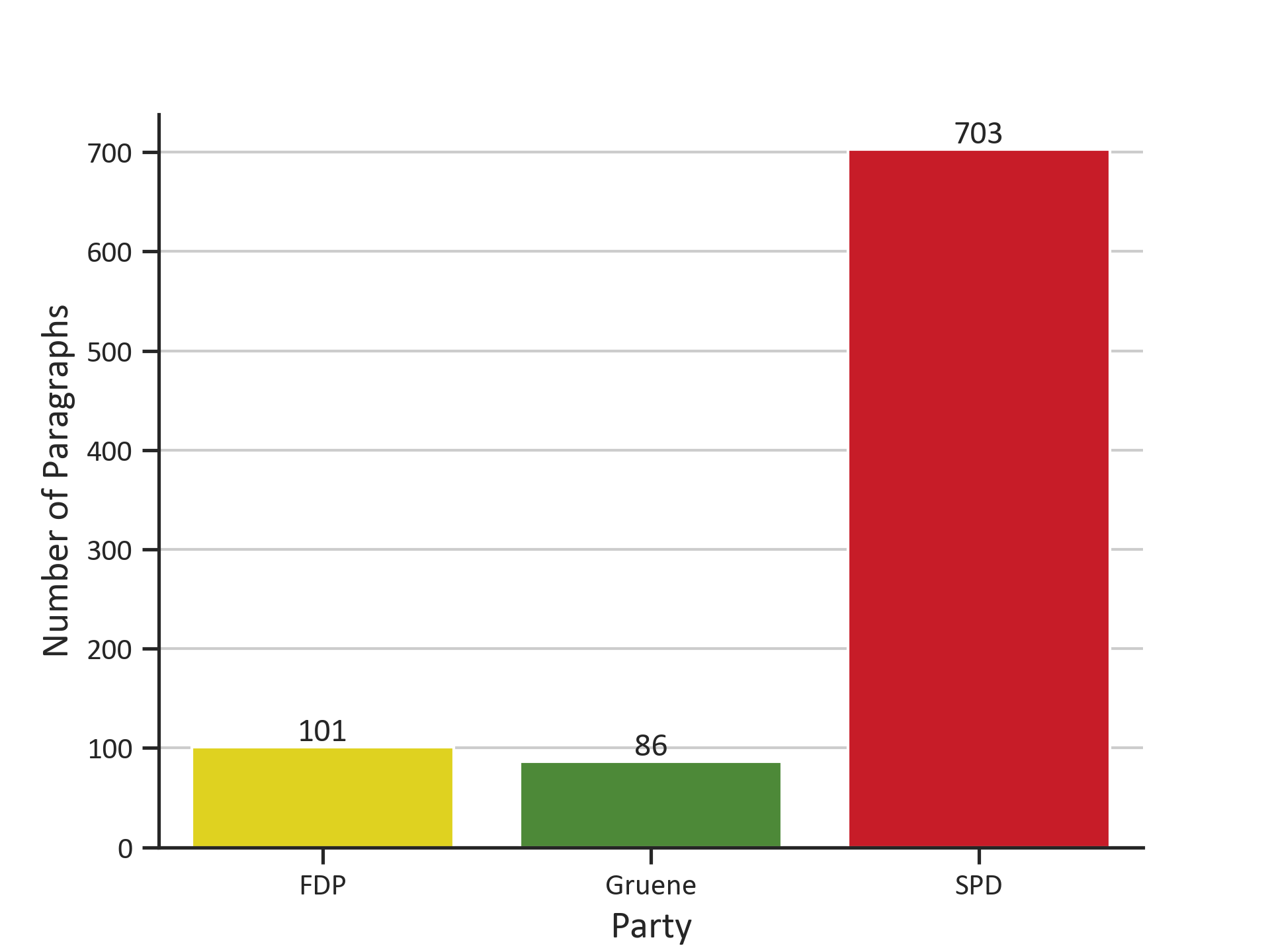}
    \caption{Classification results for paragraphs in coalition agreement (second model)}
    \label{fig:results_coal}
\end{figure}

Lastly, we apply the three-class model to the coalition agreement. Figure \ref{fig:results_coal} shows that the model attributes almost 80\% of all paragraphs to the SPD. This could be interpreted such that the SPD emerged as the winner of the negotiations. However, this result also reflects the low recall of SPD (Table \ref{tab:fscores_ampel}), where the model wrongly tends to classify a paragraph as SPD. Yet, close reading showed that the model’s certainty (softmax) was quite high (\textgreater 99\%) for numerous paragraphs, even if they could sensibly be attributed to multiple parties (e.g., in the case of minimum wage and unemployment benefits). Paragraphs with a low certainty were fairly infrequent, and mostly composed of language that is not policy critical.

Finally, keeping in mind the parties share of votes, we would have expected the Greens' proportion to be larger than that of the FDP. Instead, the latter slightly outnumbers the former. More research is needed to disseminate the roles of the smaller parties in the coalition agreement (e.g., binary classification), and also regarding an explanation of the model’s decisions.


\bibliographystyle{apalike}
\bibliography{biblio}

\end{document}